\documentclass[twocolumn]{article}
\usepackage[latin9]{inputenc}
\usepackage{array}
\usepackage{multirow}
\usepackage{amsmath}
\usepackage{amssymb}
\usepackage{graphicx}
\usepackage[unicode=true,
 bookmarks=false,
 breaklinks=false,pdfborder={0 0 1},backref=false,colorlinks=false]
 {hyperref}

\makeatletter

\providecommand{\tabularnewline}{\\}

 \usepackage{spconf}

\makeatother

\begin{document}

\title{Feature Sampling Strategies for Action Recognition}

\name{Youjie Zhou, Hongkai Yu and Song Wang}

\address{Department of Computer Science and Engineering, University of South
Carolina\\
\href{mailto:zhou42@email.sc.edu}{zhou42@email.sc.edu}, \href{mailto:yu55@email.sc.edu}{yu55@email.sc.edu},
\href{mailto:songwang@cec.sc.edu}{songwang@cec.sc.edu}}
\maketitle
\begin{abstract}
Although dense local spatial-temporal features with bag-of-features
representation achieve state-of-the-art performance for action recognition,
the huge feature number and feature size prevent current methods from
scaling up to real size problems. In this work, we investigate different
types of feature sampling strategies for action recognition, namely
dense sampling, uniformly random sampling and selective sampling.
We propose two effective selective sampling methods using object proposal
techniques. Experiments conducted on a large video dataset show that
we are able to achieve better average recognition accuracy using $25\%$
less features, through one of proposed selective sampling methods,
and even remain comparable accuracy while discarding $70\%$ features.
\end{abstract}


\section{Introduction}

Given the popularity of social media, it becomes much easier to collect
a large number of videos from Internet for human action recognition.
Effective video representation is required for recognizing human actions
and understanding video content in such rapidly increasing unstructured
data.

By far, the commonly used video representation for action recognition
has been the bag-of-words (BoW) model \cite{Peng2014a}. The basic
idea is summarizing/encoding local spatial-temporal features in a
video as a simple vector. Among local features, dense trajectory (DT)
\cite{Wang2011} and its improved variant (iDT) \cite{Wang2013} provide
state-of-the-art results on most action datasets \cite{Wang2013}.
The main idea is to construct trajectories by tracking densely sampled
feature points in frames, and compute multiple descriptors along the
trajectories.

Despite their success, DT and iDT can produce huge number of local
features, e.x., for a low resolution video in $320\times204$ with
$175$ frames, they can generate $\sim52$ Mb of features \cite{Sapienza2014}.
It is difficult to store and manipulate such dense features for large
datasets with thousands of high resolution videos, especially for
real-time applications.

Existing work focus on reducing the total number of trajectory features
through uniformly random sampling at the cost of minor reduction in
recognition accuracy. \cite{Shi2013} proposed a part model by which
they are able to randomly sample features at lower image scales in
an efficient way. \cite{Kantorov2014} interpolated trajectories using
uniformly distributed nearby feature points. \cite{Sapienza2014}
investigated the influence of random sampling on recognition accuracy
in several large scale datasets. However, intuitively, features extracted
around informative regions, such as human arms in hands waving, should
be more useful in action classification than features extracted on
the background. \cite{Mathe2012,Vig2012} proposed selective sampling
strategies on dense trajectory features based on saliency maps, produced
by modeling human eye movement when viewing videos. They are able
to achieve better recognition results with selectively sampled features.
However, it is impractical to obtain eye movement data for large datasets.

\begin{figure}
\begin{centering}
\includegraphics[width=1\columnwidth]{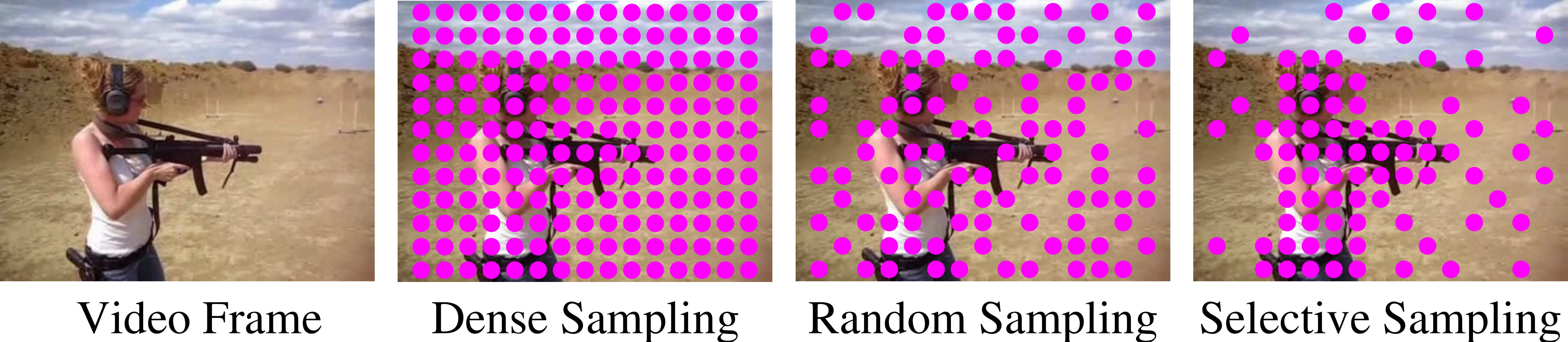}
\par\end{centering}

\caption{\label{fig:intro}Different feature sampling methods for action recognition.}
\end{figure}

In this work, we investigate several feature sampling strategies for
action recognition, as illustrated in Fig.\,\ref{fig:intro}, and
propose two data driven selective feature sampling methods. Inspired
by the success of applying object proposal techniques in efficient
saliency detection \cite{Li2013}, we construct saliency maps using
one recent object proposal method, EdgeBox \cite{Zitnick2014,Hosang2014},
and selectively sample dense trajectory features for action recognition.
We further extend EdgeBox to produce proposals and construct saliency
maps for objects with motion of interests. More effective features
can be sampled then for action classification. We evaluated a few
feature sampling methods on a publicly available datasets, and show
that proposed motion object proposal based selective sampling method
is able to achieve better accuracy using $25\%$ less features than
using the full feature set.

The remaining of this paper is organized as follows: first we give
a brief introduction about the DT/iDT features and other components
in our action classification framework, then three different feature
sampling methods are described. Finally, we discuss experimental results
on a large video dataset.

\begin{figure*}
\begin{centering}
\includegraphics[width=1\textwidth]{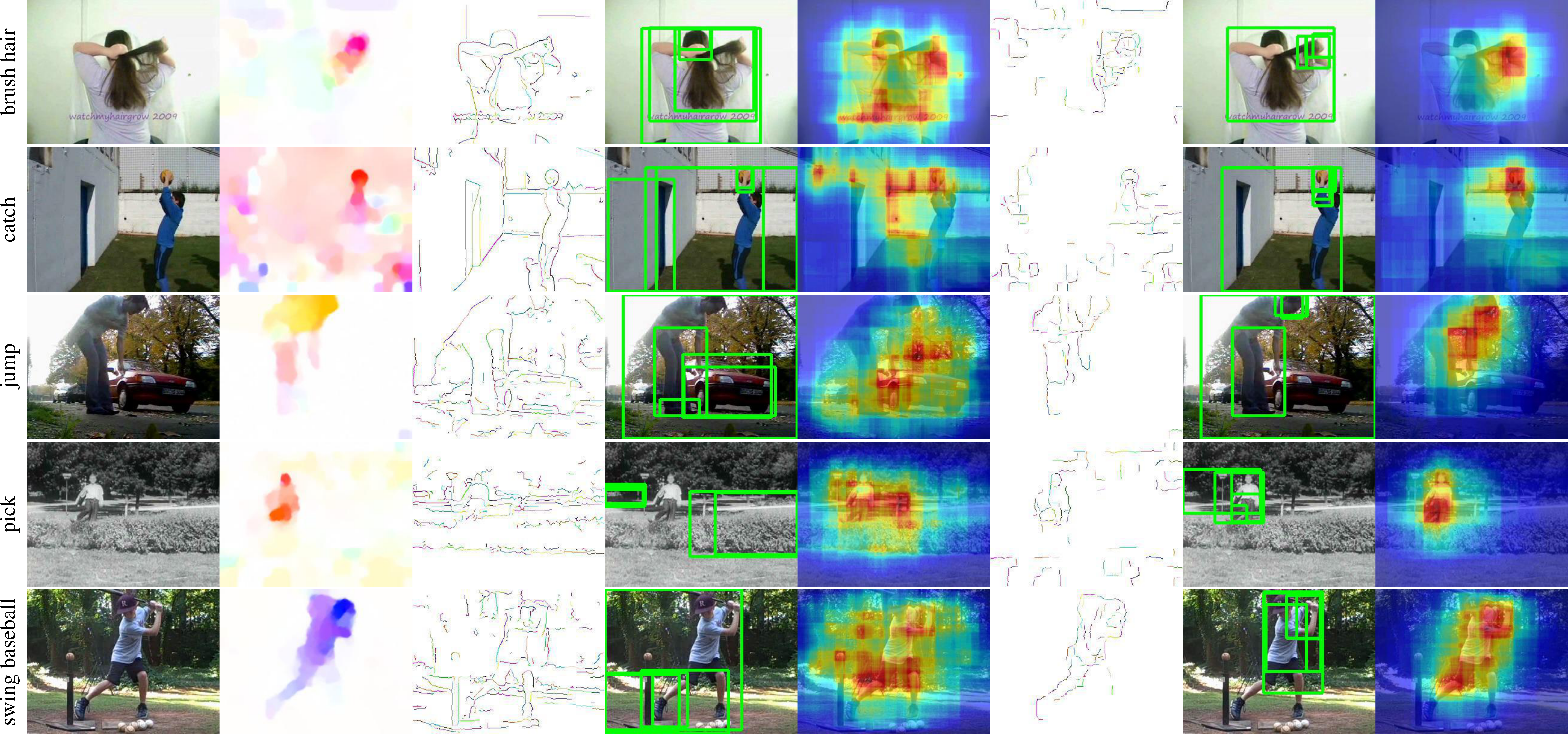}
\par\end{centering}

\caption{\label{fig:box}Illustration of selective sampling methods via object
proposal algorithms. From left to right, the original video frame,
dense optical flow field, estimated object boundaries, top $5$ scoring
boxes generated by EdgeBox, saliency map constructed using EdgeBox
proposals, estimated motion boundaries, top $5$ scoring boxes generated
by FusionEdgeBox, saliency map constructed using FusionEdgeBox.}
\end{figure*}

\section{Dense Trajectory Features}

The DT algorithm \cite{Wang2011} represents a video data by dense
trajectories, together with appearance and motion features extracted
around trajectories. On each video frame, feature points are densely
sampled using a grid with a spacing of $5$ pixels for $8$ spatial
scales spaced by a factor of $1/\sqrt{2}$, as illustrated in the
second column of Fig.\,\ref{fig:intro}. Then trajectories are constructed
by tracking feature points in the video based on dense optical flows
\cite{Farneback2003}. The default length of a trajectory is $15$,
i.e., tracking feature points in $15$ consecutive frames. The iDT
algorithm \cite{Wang2013} further enhances the trajectory construction
by eliminating background motions caused by the camera movement.

For each trajectory, $5$ types of descriptors are extracted: 1) the
shape of the trajectory encodes local motion patterns, which is described
by a sequence of displacement vectors on both x- and y-directions;
2) HOG, histogram of oriented gradients \cite{Dalal2005}, captures
appearance information, which is computed in a $32\times32\times15$
spatio-temporal volume surrounding the trajectory; 3) HOF, histogram
of optical flow \cite{Dalal2006}, focuses on local motion information,
which is computed in the same spatio-temporal volume as in HOG; 4+5)
MBHx and MBHy, motion boundary histograms \cite{Dalal2006}, are computed
separately for the horizontal and vertical gradients of the optical
flow. Both HOG, HOF and MBH are normalized appropriately.

To encode descriptors/features, we use Fisher vector \cite{Perronnin2010}
as in \cite{Wang2013}. For each feature, we first reduce its dimensionality
by a factor of two using Principal Component Analysis (PCA). Then
a codebook of size $256$ is formed by the Gaussian Mixture Model
(GMM) algorithm on a random selection of $256,000$ features from
the training set. To combine different types of features, we simply
concatenate their $l_{2}$ normalized Fisher vectors.

For classification, we apply a linear SVM provided by LIBSVM \cite{Chang2011},
and one-over-rest approach is used for multi-class classification.
In all experiments, we fix $C=100$ in SVM as suggested in \cite{Wang2013}.

\section{Feature Sampling Strategies}

In the following, we describe three feature sampling methods, that
are different from using all trajectories and related features computed
on dense grids as in the DT/iDT algorithms. All three methods can
derive a sampling probability for each trajectory feature to measure
whether it will be sampled or not, denoted by $\sigma$. For example,
$\sigma=0.8$ means we sample trajectory features with probability
greater or equal to $0.8$ for action recognition.

\subsection{Uniformly Random Sampling}

Following previous work \cite{Shi2013,Sapienza2014}, we simply sample
dense trajectory features in a random and uniform way. The sampling
probability, $\sigma$, for each trajectory is the same. In experiments,
we randomly sample $80\%$, $60\%$, $40\%$ and $30\%$ of trajectory
features, and report their action recognition accuracies respectively.

\subsection{Selective Sampling via Object Proposal}

EdgeBox \cite{Zitnick2014} is one of efficient object proposal algorithms
\cite{Hosang2014} published recently. We utilize it to construct
saliency map on each video frame, and sample trajectory features with
respect to computed saliency values.

In EdgeBox, given a video frame, object boundaries are estimated via
structured decision forests \cite{Dollar2013}, and object contours
are formed by grouping detected boundaries with similar orientations.
In order to determine how likely a bounding box contains objects of
interests, a simple but effective objectiveness score $s_{\textrm{obj}}$
was proposed, based on the number of contours that are wholly enclosed
by the box. We allow at most $10,000$ boxes in different sizes and
aspect ratios to be examined for a frame. Fig.\,\ref{fig:box} illustrates
estimated object boundaries and top $5$ scoring boxes generated by
EdgeBox in the third and forth columns respectively.

Given thousands of object proposal boxes, on a video frame, we construct
a saliency map through a pixel voting procedure. Each object proposal
box is considered as a vote for all pixels located inside it. We normalize
all pixel votes into $\left[0,1\right]$ to form a saliency probability
distribution. Saliency map examples are illustrated in the fifth column
of Fig.\,\ref{fig:box}. Warmer colors indicate higher saliency probabilities.

Based on constructed saliency maps of a video, we are able to selectively
sample trajectories and related features. If the saliency probability
of the starting pixel of a trajectory is higher than a predefined
sampling probability $\sigma$, the trajectory and related features
will be sampled. In experiments, we report recognition accuracies
for $\sigma$ with $0.2$, $0.4$ and $0.6$ respectively.

\subsection{Selective Sampling via Motion Object Proposal}

Although by stacking boxes generated via EdgeBox are able to highlight
regions in a frame with saliency objects, constructed saliency map
may not be suitable for sampling features for action recognition.
For example, in the last row of Fig.\,\ref{fig:box}, the optical
flow field (second column) clearly indicates the region with motion
of interests for action recognition is located around actor's head
and arms, while top scoring boxes and constructed saliency map via
EdgeBox incorrectly focus on actor's legs. Thus, in order to incorporate
with motion information, we propose a motion object proposal method,
named FusionEdgeBox, where a fused objectiveness score is measured
on both object boundaries and motion boundaries.

The fusion score function is defined as

\begin{equation}
s_{\textrm{fusion}}=\alpha s_{\textrm{obj}}+\beta s_{\textrm{motion}}
\end{equation}
where $s_{\textrm{obj}}$ is the original EdgeBox score, $s_{\textrm{motion}}$
is the proposed motion objectiveness score, and balance parameters
$\alpha$ and $\beta$. We empirically fix $\alpha=\beta=1$ for all
experiments. $s_{\textrm{motion}}$ is defined similar as $s_{\textrm{obj}}$,
i.e., based on the number of wholly enclosed contours in a box. However,
$s_{\textrm{motion}}$ utilizes contours that are grouped from motion
boundaries, which are estimated as image gradients of the optical
flow field. Motion boundary examples are shown in the sixth column
of Fig.\,\ref{fig:box}.

By applying the fusion score into the EdgeBox framework, we are able
to generate a set of proposal boxes, and construct the saliency map
for feature sampling as well. Examples of top $5$ scoring fusion
boxes and constructed saliency maps are illustrated in last two columns
of Fig.\,\ref{fig:box} respectively. Comparing with examples generated
by the original EdgeBox (shown in columns 3-5), we can see that FusionEdgeBox
is able to better explore regions with motion of interests, which
is useful for action feature sampling (verified by later experiments).

Similarly, we report recognition accuracies using sampled trajectory
features for $\sigma$ with $0.2$, $0.4$ and $0.6$ respectively.

\section{Experiments}

We have conducted experiments on one publicly available video datasets,
namely J-HMDB \cite{Jhuang2013}, which consists of $920$ videos
of $21$ different actions. These videos are selected from a larger
dataset HMDB \cite{Kuehne2011}. J-HMDB also provides annotated bounding
boxes for actors on each frame. We report the average classification
accuracy among three training/testing split settings provided by J-HMDB.

In the following, we evaluate action recognition on J-HMDB using sampled
trajectory features through different methods, and discuss their performance.
We also compare obtained accuracies with a few state-of-the-art action
recognition algorithms.

\subsection{Influence of Sampling Strategies}

In addition to three introduced feature sampling methods, to better
understanding trajectory features, we investigate the fourth sampling
method using annotated bounding boxes for actors. We sample trajectory
features, if the starting point of a trajectory locates inside an
annotation box. Similar strategy was proposed in \cite{Jhuang2013},
and we name it as GT.

\begin{figure}
\begin{centering}
\includegraphics[width=0.9\columnwidth]{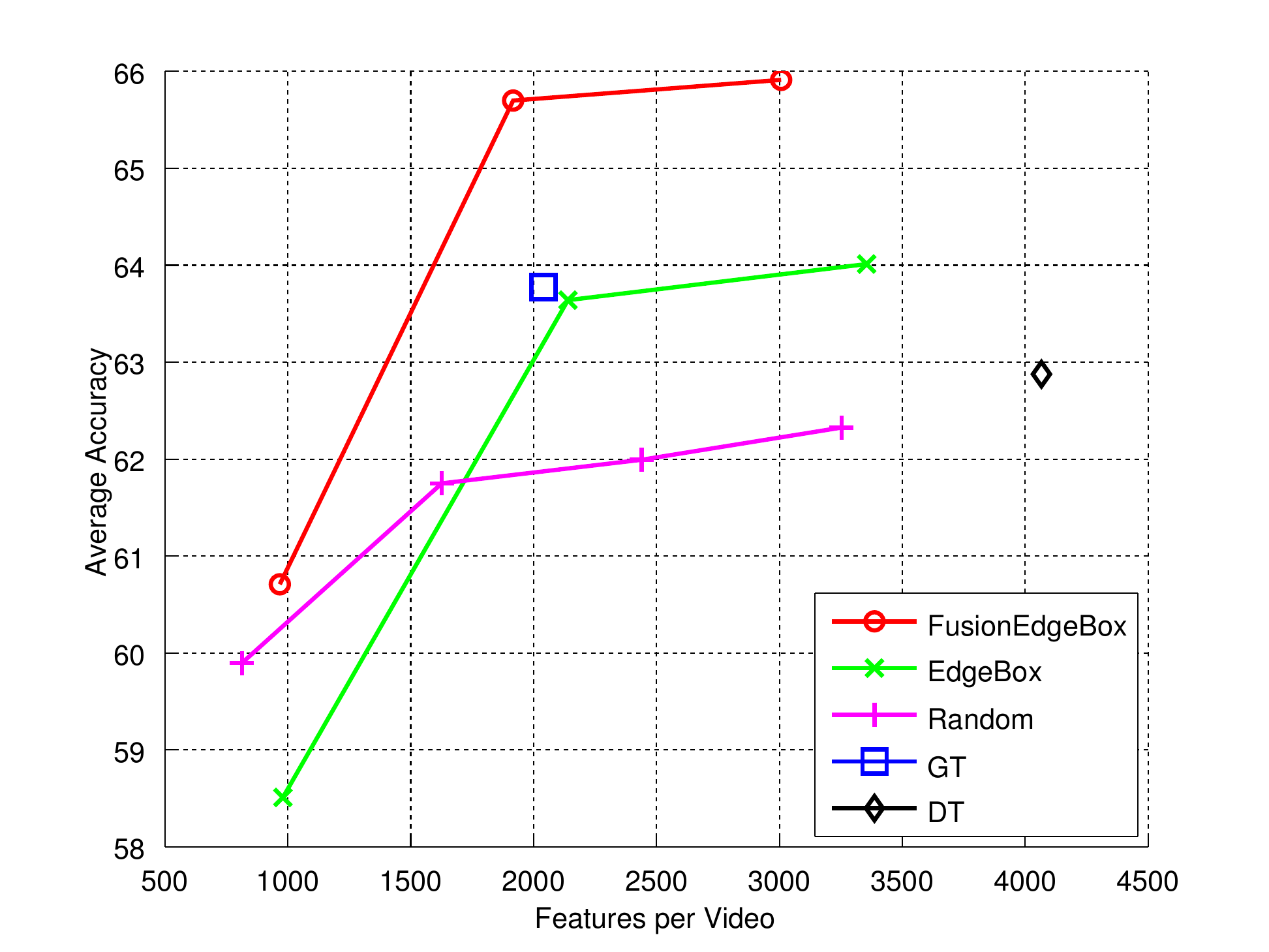}
\par\end{centering}

\caption{\label{fig:dt}Average accuracies using the DT feature.}
\end{figure}

\begin{figure}
\begin{centering}
\includegraphics[width=0.9\columnwidth]{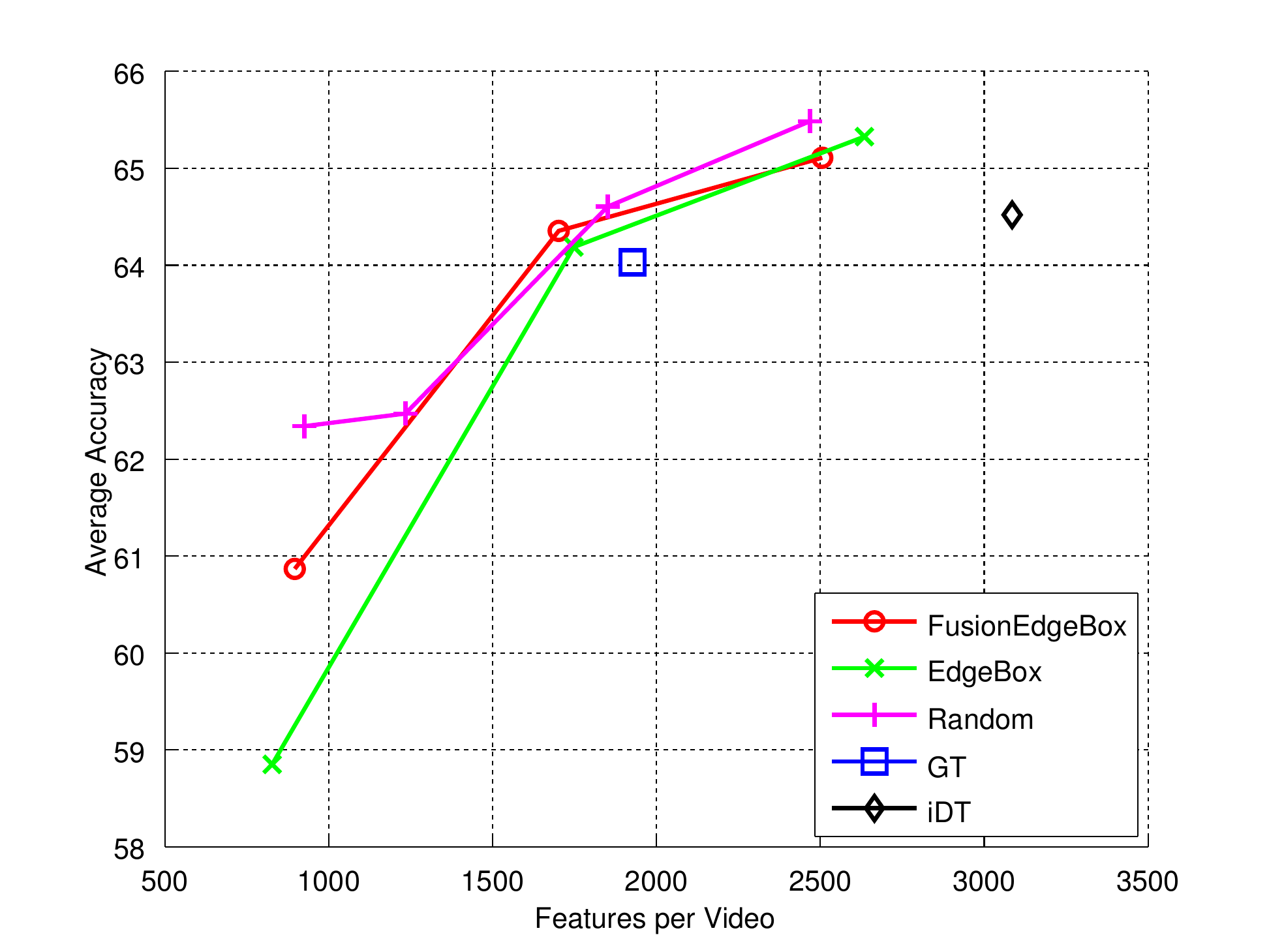}
\par\end{centering}

\caption{\label{fig:idt}Average accuracies using the iDT feature.}
\end{figure}

Figure \ref{fig:dt} and \ref{fig:idt} plot average classification
accuracies over all classes for all sampling methods under different
sampling rates, using the DT feature and iDT feature respectively.
In general, through feature sampling, we are able to achieve higher
performance than directly using all features, since noise background
features have been discarded.

Specifically, for the DT feature, we can see that: 1) trajectory features
sampled inside annotated bounding boxes, achieves higher accuracy
than using all features. Similar phenomena has been observed in \cite{Jhuang2013}
as well which indicates DT features located around human body are
more important than features extracted on other regions. 2) Selective
sampling methods achieve higher accuracies than random sampling given
similar number of sampled features. It shows that sampling DT features
from certain regions is important for action recognition, and object
proposal based strategies are able to detect these regions. 3) Proposed
selective sampling via motion object proposal outperforms other sampling
methods, even outperforms the one based on annotated bounding boxes.
It verifies that proposed FusionEdgeBox method is useful for exploring
regions of interests for action recognition.

For the iDT feature, however, different sampling method result in
similar accuracies. Random sampling outperforms others slightly, especially
when the number of sampled features is small. The reason may  be that,
by eliminating background motion caused by the camera movement, the
iDT feature is more compact and meaningful than the DT feature, e.x.,
the average number of iDT features per video is much lower than it
of DT feature. Random sampling is able to better preserve the original
iDT feature distribution than selective samplings which have quite
large sampling bias.


\subsection{Comparisons to state-of-the-arts}

\begin{table}
\begin{centering}
{\small{}}%
\begin{tabular}{|>{\raggedright}p{0.07\columnwidth}|c|c|c|}
\hline 
\multicolumn{2}{|c|}{{\small{}Method}} & {\small{}J-HMDB} & {\small{}Memory (GB)}\tabularnewline
\hline 
\hline 
\multicolumn{2}{|c|}{{\small{}Dense Trajectory \cite{Wang2011}}} & {\small{}$62.88\%$} & {\small{}$5.4$}\tabularnewline
\hline 
\multicolumn{2}{|c|}{{\small{}Improved Dense Trajectory \cite{Wang2013}}} & {\small{}$64.52\%$} & {\small{}$4.2$}\tabularnewline
\hline 
\multicolumn{2}{|c|}{{\small{}Peng }\emph{\small{}et al.}{\small{} \cite{Peng2014} w/
iDT}} & {\small{}$69.03\%${*}} & {\small{}$4.2$}\tabularnewline
\hline 
\multicolumn{2}{|c|}{{\small{}Gkioxari }\emph{\small{}et al.}{\small{} \cite{Gkioxari2014}}} & {\small{}$62.5\%$} & {\small{}-}\tabularnewline
\hline 
\hline 
\multicolumn{4}{|c|}{{\small{}Discard $20\%\sim25\%$ features}}\tabularnewline
\hline 
\multirow{3}{0.07\columnwidth}{{\small{}DT}} & {\small{}Random} & {\small{}$62.33\%$} & {\small{}$4.3$}\tabularnewline
\cline{2-4} 
 & {\small{}EdgeBox} & {\small{}$65.33\%$} & {\small{}$4.5$}\tabularnewline
\cline{2-4} 
 & {\small{}FusionEdgeBox} & \textbf{\small{}$\mathbf{65.91\%}$} & {\small{}$4.0$}\tabularnewline
\hline 
\multirow{3}{0.07\columnwidth}{{\small{}iDT}} & {\small{}Random} & {\small{}$\mathbf{65.49\%}$} & {\small{}$3.4$}\tabularnewline
\cline{2-4} 
 & {\small{}EdgeBox} & {\small{}$65.32\%$} & {\small{}$3.6$}\tabularnewline
\cline{2-4} 
 & {\small{}FusionEdgeBox} & {\small{}$65.11\%$} & {\small{}$3.5$}\tabularnewline
\hline 
\hline 
\multicolumn{4}{|c|}{{\small{}Discard $70\%\sim80\%$ features}}\tabularnewline
\hline 
\multirow{3}{0.07\columnwidth}{{\small{}DT}} & {\small{}Random} & {\small{}$59.90\%$} & {\small{}$1.1$}\tabularnewline
\cline{2-4} 
 & {\small{}EdgeBox} & {\small{}$58.51\%$} & {\small{}$1.4$}\tabularnewline
\cline{2-4} 
 & {\small{}FusionEdgeBox} & {\small{}$\mathbf{60.71\%}$} & {\small{}$1.4$}\tabularnewline
\hline 
\multirow{3}{0.07\columnwidth}{{\small{}iDT}} & {\small{}Random} & {\small{}$\mathbf{62.34\%}$} & {\small{}$1.3$}\tabularnewline
\cline{2-4} 
 & {\small{}EdgeBox} & {\small{}$58.85\%$} & {\small{}$1.2$}\tabularnewline
\cline{2-4} 
 & {\small{}FusionEdgeBox} & {\small{}$60.87\%$} & {\small{}$1.3$}\tabularnewline
\hline 
\end{tabular}
\par\end{centering}{\small \par}

\caption{\label{tab:cmp}Comparison to state-of-the-arts in terms of average
accuracy and feature size. {*} It leverages an advanced feature encoding
technique, stacked Fisher vector.}

\end{table}
Table \ref{tab:cmp} shows comparisons of feature sampling methods
in different sampling rates with the state-of-the-arts. Sampling methods
achieve better average accuracies than a few state-of-the-arts using
same classification pipeline, with $\sim20\%$ less features. It is
interesting to observe that, even discarding more than $70\%$ features,
random sampling and proposed selective sampling still are able to
remain comparable performance.

\section{Conclusions}

In this work, we focus on feature sampling strategies for action recognition
in videos. Dense trajectory features are utilized to represent videos.
Two types of sampling strategies are investigated, namely uniformly
random sampling and selective sampling. We propose to use object proposal
techniques to construct saliency maps for video frames, and use them
to guide the selective feature sampling process. We also propose a
motion object proposal method that incorporate object motion information
into object proposal framework. Experiments conducted on a large video
dataset indicate that sampling based methods are able to achieve better
recognition accuracy using $25\%$ less features through one of proposed
selective feature sampling method, and even remain comparable accuracy
with discarding $70\%$ features.

\bibliographystyle{icip}
\bibliography{ActFea-ICIP2015}

\end{document}